\documentclass[letterpaper,final]{article}
\usepackage{nips_ml4h_2017}

\usepackage[utf8]{inputenc} 
\usepackage[T1]{fontenc}    
\usepackage{hyperref}       
\usepackage{url}            
\usepackage{booktabs}       
\usepackage{microtype}      
\usepackage{nicefrac}      
\usepackage{authblk}
\usepackage{enumitem}

\usepackage{amsmath,amssymb,amsfonts}
\newcommand{\ts}{\textstyle}

\usepackage{color}

\usepackage{natbib}
\setcitestyle{numbers,open={[},close={]}}

\usepackage{graphicx}
\graphicspath{{figures/}}
\usepackage{xcolor}
\usepackage[font=small]{caption}
\usepackage{chngcntr}

\usepackage{float}
\usepackage{algorithm}
\usepackage[noend]{algpseudocode}
\algrenewcommand\algorithmicfunction{\textbf{def}}
\algrenewcommand\algorithmicrequire{\textbf{Input:}}
\algrenewcommand\algorithmicensure{\textbf{Output:}}

\usepackage{listings}
\makeatletter
\newcommand{\verbatimfont}[1]{\renewcommand{\verbatim@font}{\ttfamily#1}}
\makeatother

\title{Prediction-Constrained Topic Models
for Antidepressant Recommendation}
\author[1]{\textbf{Michael C. Hughes}}
\author[2]{\textbf{Gabriel Hope}}
\author[3]{\textbf{Leah Weiner}}
\author[4]{\\ \textbf{Thomas H. McCoy, M.D.}}
\author[4]{\textbf{Roy H. Perlis, M.D.}}
\author[2]{\textbf{Erik B. Sudderth}}
\author[1]{\textbf{Finale Doshi-Velez}}
\affil[1]{School of Engineering and Applied Sciences, Harvard University}
\affil[2]{School of Information \& Computer Sciences, Univ. of California, Irvine}
\affil[3]{Dept. of Computer Science, Brown University}
\affil[4]{Massachusetts General Hospital}

\begin{document}
\maketitle


\begin{abstract}
Supervisory signals can help topic models discover low-dimensional data
representations that are more interpretable for clinical tasks. We propose a
framework for training supervised latent Dirichlet allocation that balances
two goals: faithful generative explanations of high-dimensional data and
accurate prediction of associated class labels. Existing approaches fail to
balance these goals by not properly handling a fundamental asymmetry: the
intended task is always predicting labels from data, not data from
labels.
Our new prediction-constrained objective trains models that predict labels
from heldout data well while also producing good generative likelihoods and
interpretable topic-word parameters.  In a case study on predicting depression
medications from electronic health records, we demonstrate improved
recommendations compared to previous supervised topic models and high-dimensional logistic regression from words alone.

\end{abstract}

\section{Introduction}
Patient history in electronic health records (EHR) can be represented as counts of predefined concepts like procedures, labs, and medications.
For such datasets, topic models such as \emph{latent Dirichlet allocation}~(LDA,~\citep{blei2003lda}) are popular
 for extracting insightful low-dimensional structure for clinicians~\citep{paul2014discovering,ghassemi2014unfolding}.
A natural goal is to use such low-dimensional representations as features for a specific \emph{supervised prediction}, such as recommending drugs to a patient with depression. Many general-purpose efforts have attempted to train \emph{supervised topic models} \citep{zhu2012medlda,lacoste2009disclda,chen2015bplda}, 
including the well-known supervised Latent Dirichlet Allocation~(sLDA,
\citep{blei2007sLDA}).

However, a recent survey of healthcare prediction tasks~\citep{halpern2012comparison} finds
that many of these approaches have little benefit, if any, over standard unsupervised LDA for heldout predictions.
In this work, we expose and correct several deficiencies in these previous formulations of supervised topic models.  We introduce a learning objective that directly enforces the intuitive goal of representing the data in a way that enables accurate downstream predictions.  Our objective acknowledges the inherent asymmetry of prediction tasks: 
clinicians want to predict medication outcomes given medical records, not medical records given outcomes.  Approaches like sLDA that optimize the \emph{joint} likelihood of labels and words ignore this crucial asymmetry.  
Our new \emph{prediction-constrained} (PC) objective for training latent variable models 
allows practitioners to effectively balance explaining abundant count data while ensuring high-quality predictions of labels from this data.
We hope achieving strong gains in this predictive framework will pave the way for causal latent variable models for drug recommendation.

\vspace*{-5pt}
\section{Limitations of Existing Topic Models}
\vspace*{-5pt}
\paragraph{Supervised LDA.} The sLDA model learns from a collection of $D$ documents. Each document $d$ is represented by counts of $V$ discrete words, $x_d \in \mathbb{Z}_+^V$. In the EHR context, these are often ICD-9 or ICD-10 codes, such as ``F33.0: Major depressive disorder, recurrent, mild''. For our supervised case, each document (patient) $d$ also has a binary label $y_d \in \{0,1\}$, indicating whether a medication was successful. Both words $x_d$ and label $y_d$ are generated by a document-specific mixture of $K$ topics:
\begin{align}
x_d | \pi_d, \phi &\sim \mbox{Mult}( x_d \mid 
        \textstyle \sum_{k=1}^K \pi_{dk} \phi_{k}, N_d ),
        \quad
y_d | \pi_d, \eta \sim \mbox{Bern}( y_d \mid \sigma(
        \textstyle \sum_{k=1}^K \pi_{dk} \eta_{k} ) ).
\label{eq:generative_model}        
\end{align}
The key latent variable is $\pi_d$, the document-topic probability vector, with prior $\pi_d \sim \mbox{Dir}(\alpha)$.
The trainable parameters are topic-word probabilities $\phi_{k}$ and 
regression weights $\eta_k$ (we fix $\alpha$ for simplicity).
Let $\sigma(z) = (1+e^{-z})^{-1}$ be the logit function, and $N_d$ the observed size of document~$d$.

There are a host of objectives and inference methods for supervised LDA, including~\citep{blei2007sLDA, wang2009simultaneous,zhang2014howToSuperviseTopicModels,zhu2012medlda}. A key contribution of this work is identifying a fundamental shortcoming of all these objectives: they do not train topic models that are also effective at label prediction.  This myriad of methods, and their shortcomings, arise because of model mispecification.  If our count data truly came from a topic model, and those topics truly led to good label predictions, then even \emph{unsupervised} topic models would do well.  Trouble arises when we desire the dimensionality reduction provided by a topic model for interpretability or efficiency, but the data were not produced by the LDA generative process.

\vspace*{-10pt}
\paragraph{Limitations of Joint Bayesian or Maximum-Likelihood Training of the sLDA Model.}
Supervised LDA \citep{blei2007sLDA} and related work \citep{wang2009simultaneous,wang2014spectral,ren2017spectral} assumes a graphical model in which the target label $y_d$ can be viewed as yet another output of document-topic probabilities $\pi_d$.
When the number of counts in $x_d$ is significantly larger than the cardinality of $y_d$, as is typical in practice, the likelihood associated with $x_d$ will be much larger in magnitude than the likelihood associated with $y_d$.  That is, the \emph{correct} application of Bayesian inference within this model will essentially \emph{ignore} the task of predicting the target $y_d$. Thus, \citep{halpern2012comparison} finds that for large $K$, sLDA is no better than LDA. 

\vspace*{-10pt}
\paragraph{Limitations of Label Replication.}
The Power-sLDA approach of \citet{zhang2014howToSuperviseTopicModels} suggests improving  sLDA's predictions by artificially replicating the label $y_d$ multiple times. 
Standard Bayesian methods use both data $x_d$ and replicated labels $y_d$ to infer the document-topic probabilities $\pi_d$ while training.
However, the predictive posterior $p(\pi_d \mid x_d)$ may be very different from the training posterior $p(\pi_d \mid x_d, y_d)$.
Put another way, label replication strengthens the connection between $\pi_d$ and $y_d$, but it does not strengthen the task we care about: prediction of $y_d$ from $x_d$ alone.
Figure~\ref{fig:results_toy_haystack} in the appendix demonstrates this issue: regardless of the replication level, when the model is misspecified Power-sLDA fails to find topics that are good for predicting $y_d$ from $x_d$.

\vspace*{-10pt}
\paragraph{Other popular sLDA objectives reduce to label replication.}
Posterior regularization (PR)-based methods~\citep{ganchev2010posteriorconstraints,gracca2008posteriorconstraints} enforce explicit performance constraints on the posterior.  The MedLDA approach of \citet{zhu2012medlda, zhu2013gibbsmaxmargin,zhu2014regbayes} is instead derived from a maximum entropy discrimination framework, and uses a hinge loss to penalize errors in the prediction of $y_d$.
In extended derivations in our longer tech report~\citep{hughes2017pc}, we show that both MedLDA and PR training objectives can be written as instances of label replication, and thus inherit Power-sLDA's failure to generalize well.

\vspace*{-10pt}
\paragraph{Limitations of Fully Discriminative Learning.}
Unlike the above approaches, backpropagation supervised LDA~(BP-sLDA,~\citep{chen2015bplda}) focuses \emph{entirely} on the prediction of $y_d$. 
\citet{chen2015bplda} do handle the direct prediction of $y_d$ from $x_d$, but no term in their objective forces topics to accurately model the data $x_d$ at all. Our objective can be seen as a principled generalization that balances the explanation of data $x_d$ (which \cite{chen2015bplda} ignores) and prediction of targets $y_d$. Our improved generative modeling leads to more interpretable topic-word distributions.

\vspace*{-2pt}
\section{Prediction-Constrained sLDA}
\vspace*{-2pt}
We propose a novel, \emph{prediction-constrained} (PC) objective that explicitly encodes the asymmetry of the discriminative label prediction task.
In particular, we ensure that topics learned during joint training can also be used to make accurate predictions about $y$ given $x$, by solving:
\begin{align}
\label{eq:our_objective}
\underset{\phi,\eta}{\min}\;
    - \Big[ \ts \sum_{d=1}^D
        \log p(x_d \mid \phi , \alpha ) 
    \Big]
\mbox{~subject to~} 
    - \ts \sum_{d=1}^D \log p( y_d \mid x_d , \phi , \eta , \alpha ) \leq \epsilon.
\end{align}
The scalar $\epsilon$ is the highest aggregate loss we are willing to tolerate. There are many variations on this theme; for example, one could instead use a hinge loss as in \citet{zhu2012medlda}.  The structure of Eq.~\eqref{eq:our_objective} matches the goals of a domain expert who wishes to explain as much of the data $x$ as possible, while still making sufficiently accurate predictions. We recommend adding standard regularization terms (log priors for $\phi$ and $\eta$), though we leave these out to keep notation focused on our contributions.

Applying the Karush-Kuhn-Tucker conditions, Eq.~\eqref{eq:our_objective} becomes an equivalent unconstrained problem:
\begin{align}
\underset{\phi,\eta}{\min}
&
- \ts \sum_{d=1}^D [
     \log p(x_d | \phi, \alpha)
    + \lambda_{\epsilon} \log p( y_d | x_d , \phi , \eta, \alpha)
    ]
\label{eq:unconstrained_objective}
\end{align}
For any prediction tolerance $\epsilon$, there exists a scalar multiplier $\lambda_\epsilon > 0$ such that the optimum of Eq.~\eqref{eq:our_objective} is a minimizer of Eq.~\eqref{eq:unconstrained_objective}.
The relationship between $\lambda_\epsilon$ and $\epsilon$ is monotonic but has no analytic form. We must search over one-dimensional penalties $\lambda_\epsilon$ for an appropriate value.  

While our PC objective is superficially similar to Power-sLDA~\citep{zhang2014howToSuperviseTopicModels} and MedLDA~\citep{zhu2012medlda}, it is distinct: the multiplier $\lambda_\epsilon$ rescales the log-posterior $\log p(y_d \mid x_d)$, while label-replication rescales the log-likelihood $\log p(y_d \mid \pi_d)$.  By ``replicating'' the \emph{entire} posterior, rather than just the link between latent and target variables, our PC objective achieves the asymmetric goal of predicting $y_d$ from $x_d$ alone.

Computing $p(x_d \mid \phi )$ and $p( y_d \mid x_d , \phi , \eta)$ requires marginalizing $\pi_d$ over the simplex. However, these integrals are intractable. To gain traction, we first contemplate \emph{instantiating} $\pi_d$:
\begin{align}
\ts \min_{\pi, \phi, \eta} 
- \ts \sum_{d=1}^D  [&
     \log p(\pi_d | \alpha) + \log p(x_d | \pi_d, \phi)
     + \lambda_{\epsilon} \log p(y_d | \pi_d, \eta) ]
\label{eq:label_rep_objective}
\end{align}
As discussed above, solutions to this objective would lead to weighted \emph{joint} training and its symmetry problems. Since we wish to train under the same asymmetric conditions needed at test time, where we have $x_d$ but not $y_d$, we instead \emph{fix} $\pi_d$ to a deterministic mapping of the words $x_d$ to the topic simplex. 
Specifically, we fix to the \emph{maximum a posteriori} (MAP) solution
$\pi_d = \mbox{argmax}_{\pi_d \in \Delta^K} \log p(\pi_d | x_d, \phi, \alpha)$, which we write as an embedding:
$\pi_d \gets \mbox{MAP}_{\phi,\alpha}( x_d )$.

Our chosen embedding can be seen as a feasible
approximation to the posterior $p( \pi_d | x_d , \phi , \alpha )$.
This choice respects the need to use the same embedding of observed words $x_d$ into low-dimensional $\pi_d$ in both training and test scenarios. We can now write our tractable training objective for PC-sLDA:
\begin{align}
\label{eq:our_differentiable_obj}
- \ts \sum_{d=1}^D [
    &\log p(\mbox{MAP}_{\phi,\alpha}(x_d) | \alpha)
    + \log p(x_d | \mbox{MAP}_{\phi,\alpha}( x_d) , \phi )
    + \lambda_{\epsilon} \log p( y_d | \mbox{MAP}_{\phi,\alpha}(x_d), \eta )
    ]
\end{align}
While this objective is similar to BP-sLDA~\citep{chen2015bplda}, the key difference is that
our method \emph{balances} the generative and discriminative terms via the multiplier $\lambda_\epsilon$. In contrast, \citet{chen2015bplda} consider only fully unsupervised (labels $y$ are ignored) or fully supervised (the distribution of $x$ is ignored) cases.

\paragraph{MAP via Exponentiated Gradient.}
The document-topic MAP problem for unsupervised LDA is
$\max_{\pi_d \in \Delta^{K}}
\log p(\pi_d | x_d, \phi, \alpha)$ \citep{sontag2011complexityoflda}.
It is convex for $\alpha \geq 1$ and non-convex otherwise. For the convex case, we
start from uniform probabilities and iteratively do \emph{exponentiated gradient} updates~\citep{kivinen1997exponentiated_gradient}:
\begin{align}
\mbox{init:~~}&
\pi^{0}_{d} \gets [\frac{1}{K} \ldots \frac{1}{K}],
\quad \mbox{repeat:~}
\pi^{t}_{dk} \gets \ts \frac{p^t_{dk}}{ \sum_{j=1}^{K} p^t_{dj} },
\quad
p^t_{dk} = \pi^{t-1}_{dk} \circ e^{ \nu \nabla \log p(\pi^{t-1}_{d} | x_d)}.
\label{eq:expgrad}
\end{align}
With small enough steps $\nu > 0$, exponentiated gradient converges
to the MAP solution.
We define our embedding $\mbox{MAP}_{\phi,\alpha}(x_d)$
to be the deterministic outcome of $T$ iterations of Eq.~\eqref{eq:expgrad}. $T \approx 100$ and $\nu \approx 0.005$ work well. The non-convex case can be solved similarly  after reparameterization~\citep{taddy2012topicmodelmapestimation, mackay1998basis}.


\paragraph{Learning via gradient descent.}
Our entire objective function, including the MAP estimation procedure, is fully \emph{differentiable} with respect to the parameters $\phi, \eta$. 
Thus, modern gradient descent methods like Adam may be applied to estimate $\phi, \eta$ from observed data. We have developed Python implementations using both Autograd \citep{maclaurin2015autograd} and Tensorflow \citep{tensorflow2015whitepaper}, which we will release to the public.

\vspace*{-10pt}
\paragraph{Hyperparameter selection.}
The key hyperparameter for our PC-sLDA algorithm is the multiplier $\lambda_\epsilon$.  For topic models, $\lambda_\epsilon$ typically needs to scale like the number of tokens in the average document, though it may need to be larger depending on tension between the unsupervised and supervised terms of the objective. In our experiments, we try logarithmically spaced values $\lambda_{\epsilon} \in \{10, 100, 1000, \ldots \}$ and select the best using validation data, although this requires training multiple models. This cost can be somewhat mitigated by using the final parameters at one $\lambda_\epsilon$ value to initialize the next $\lambda_\epsilon$, although this may not escape to new preferred basins of attraction in the overall non-convex objective.

\section{Antidepressant Case Study}
\label{sec:results}

We consider predicting which subset of 11 common antidepressants will be successful for a patient with major depressive disorder given a bag-of-words representation $x_d$ of the patient's electronic health record (EHR).  These are real deidentified data from tertiary care hospitals, split into 29774/3721/3722 documents (one per patient) with $V=5126$ EHR codewords (diagnoses/procedures/medicines). The appendix gives details on data preprocessing.
Our results are:

%

\begin{figure*}
\begin{tabular}{c c r}
\begin{minipage}{0.28\textwidth}
\begin{tabular}{c}
\includegraphics[width=\textwidth]{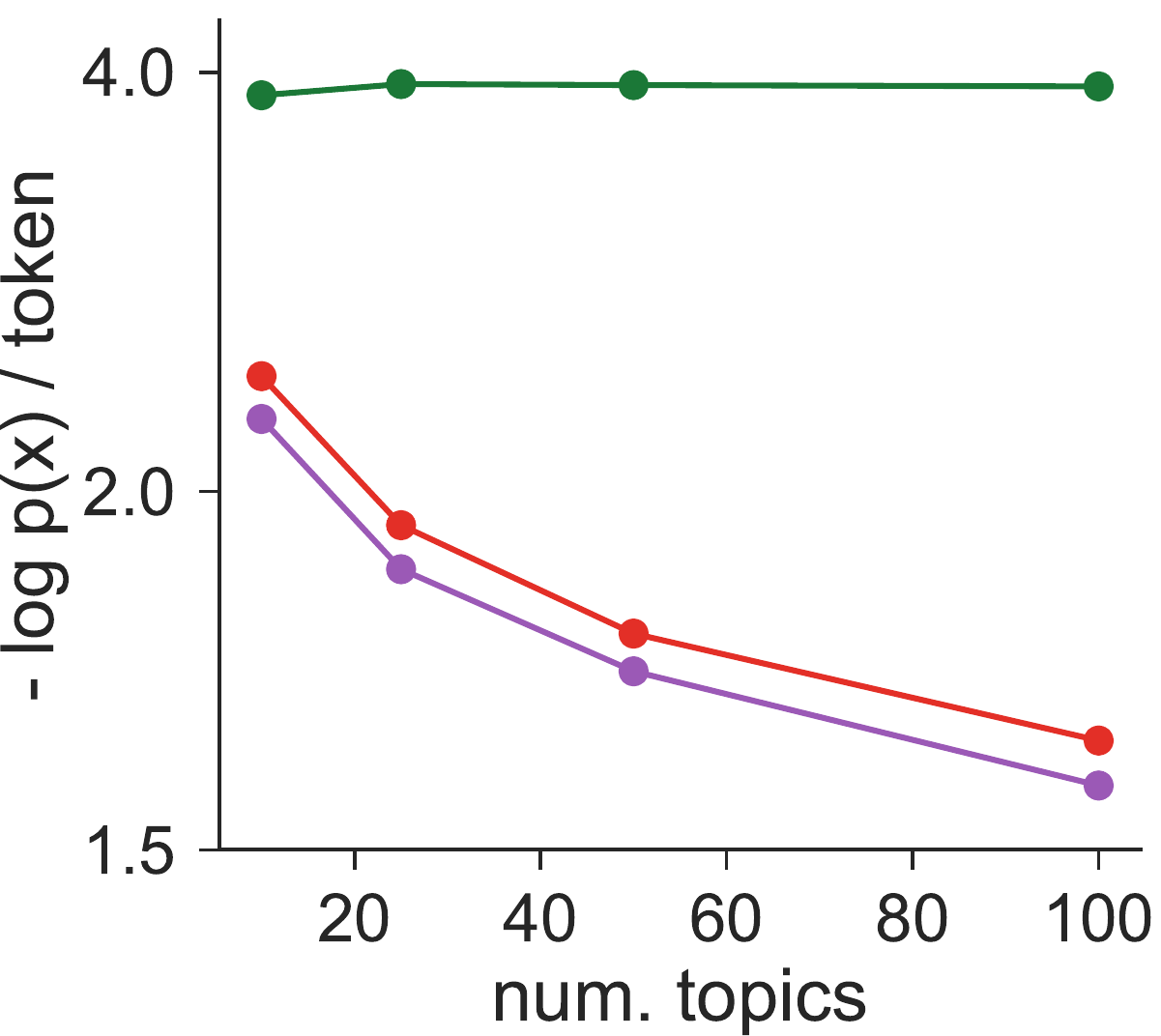}
\\
\includegraphics[width=\textwidth]{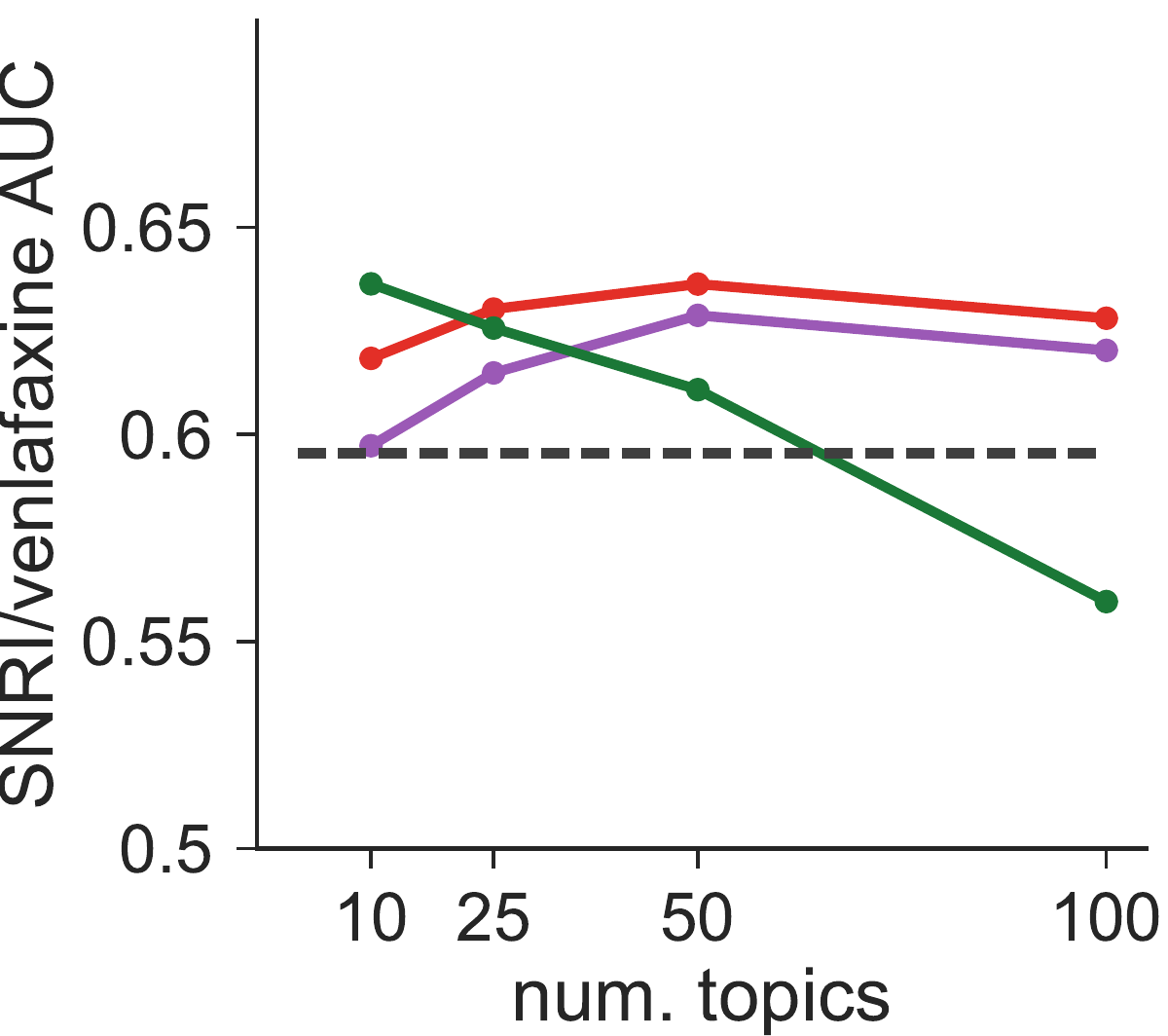}
\end{tabular}
\end{minipage}
&
\begin{minipage}{0.28\textwidth}
\begin{tabular}{c}
\includegraphics[width=\textwidth]{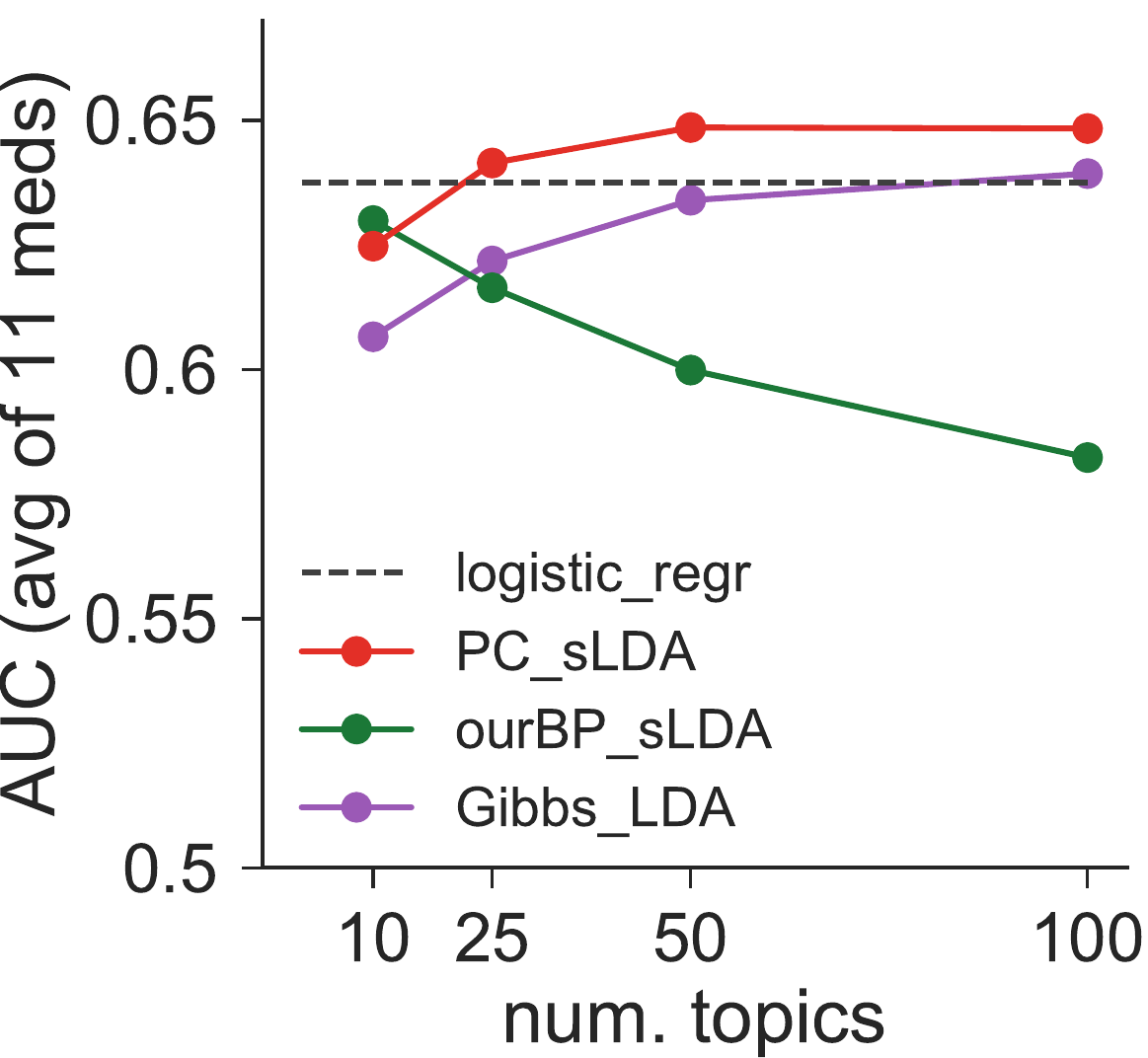}
\\
\includegraphics[width=\textwidth]{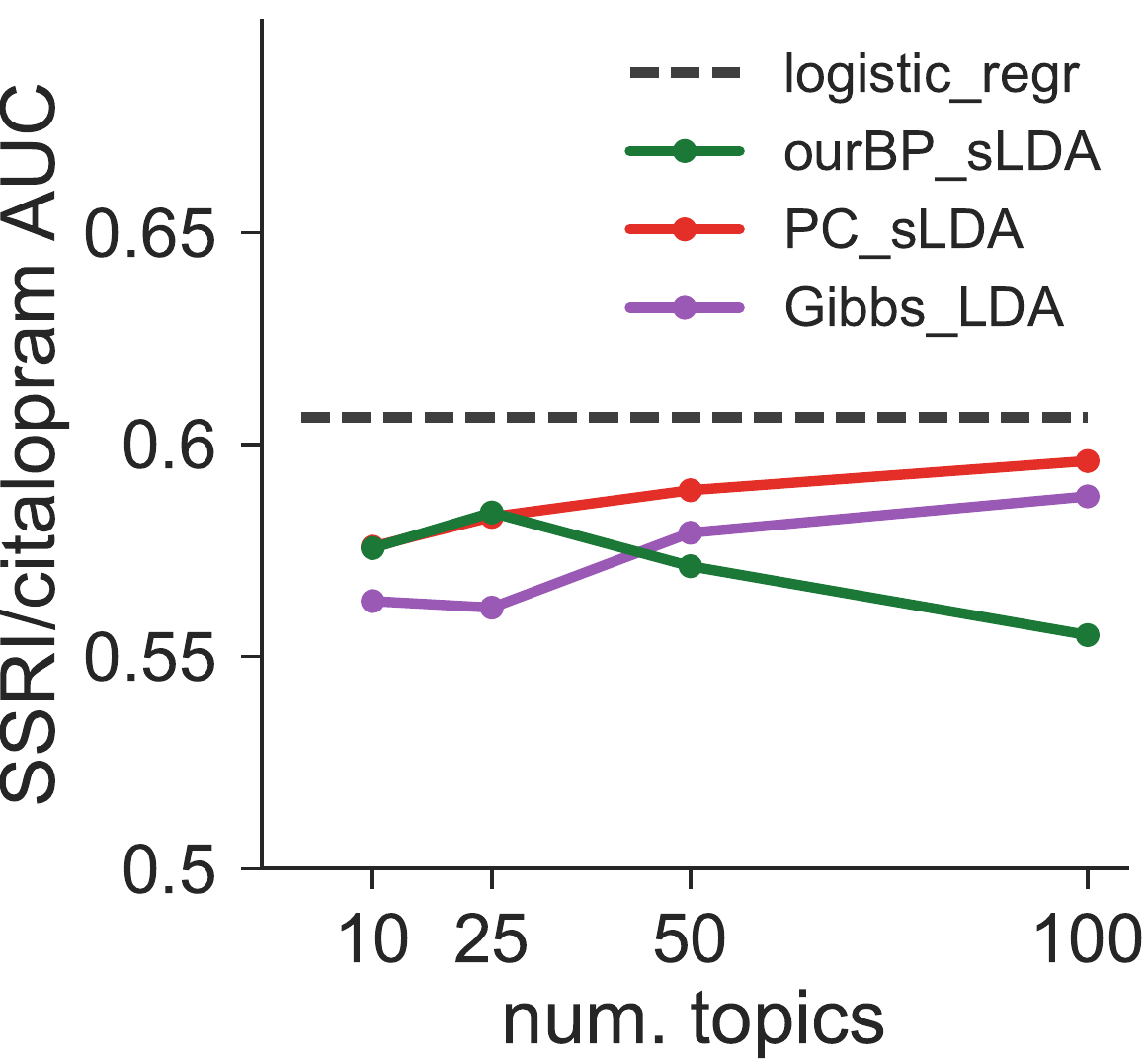}
\end{tabular}
\end{minipage}
&
\verbatimfont{\tiny}%
\begin{minipage}{0.4\textwidth}
{\small PC sLDA $\eta_{k}=$ ~+0.3}
\begin{lstlisting}
cpt  99213 0.137 office_visit>=15min
cpt  99211 0.037 office_visit>=05min
cpt  99214 0.030 office_visit>=25min
icd9  v700 0.029 routine_physical_exam
cpt  85027 0.021 complete_blood_ct_test
cpt  36415 0.021 routine_blood_collection
icd9 78079 0.015 other_malaise_&_fatigue
-----
cpt  87880 0.987 test_for_strep_throat
icd9 70583 0.982 hidradenitis_skin_cond
cpt  99403 0.963 preventive_counsel>=45min
icd9  v031 0.963 need_for_typhoid_vaccine
\end{lstlisting}
{\small Gibbs LDA $\eta_{k}=$~-0.6}
\begin{lstlisting}
cpt  99213 0.083 office_visit>=15min
cpt  99211 0.045 office_visit>=05min
icd9  v700 0.026 routine_physical
cpt  99214 0.025 office_visit>=25min
cpt  36415 0.015 complete_blood_ct_test
cpt  99212 0.015 office_visit>=10min
cpt  90658 0.013 flu_vaccine

cpt  90733 1.000 meningococcal_vaccine
cpt  90713 1.000 poliovirus_vaccine
cpt  99393 0.999 routine_physical_age5-11
cpt  90691 0.999 typhoid_vaccine
\end{lstlisting}
\end{minipage}%
\end{tabular}
\vspace{-8pt}
\caption{
Antidepressant prediction. 
\emph{Left:} Performance as number of topics increases. We show heldout negative log likelihood (generative, lower is better)
and heldout AUC (discriminative, higher is better) for the average of all 11 drugs as well as two specific drugs (one SNRI and one SSRI).
We use our own implementation of BP-sLDA for this multiple binary label prediction task. Both PC-sLDA and BP-sLDA results use runs initialized from Gibbs. While BP-sLDA exhibits severe overfitting, our PC-sLDA improves on the baseline Gibbs predictions reliably.
\emph{Right:} 
Interpretation of topic \#11 of $K=25$ for both Gibbs-LDA and our PC-sLDA initialized from Gibbs.
We show the regression coefficient $\eta_k$ for this topic when predicting patient success with citalopram.
The top list is ranked by $p(\textnormal{word}|\textnormal{topic})$; the bottom list by
$p(\textnormal{topic}|\textnormal{word})$, indicating potential \emph{anchor words}.
The original Gibbs topic is mostly about routine preventative care and vaccination. 
PC sLDA training evolves the topic to emphasize 
longer duration encounters focused on counseling, with a few unfocused terms.
}
\label{fig:results_psych}
\end{figure*}

\textbf{PC-sLDA has better label prediction.}
Overall, antidepressant recommendation is challenging even for nonlinear classifiers, so we do not expect AUC scores to be very high. However, our PC-sLDA is competitive, beating Gibbs LDA and logistic regression at average prediction across 11 drugs in Fig.~\ref{fig:results_psych} when given enough topics. BP-sLDA does well with few topics, but overfits with too many.

\textbf{PC-sLDA recovers better heldout data likelihoods than BP-sLDA.}
Fig.~\ref{fig:results_psych} shows trends in negative log likelihood on heldout data (lower is better). 
As expected, unsupervised Gibbs-LDA consistently achieves the best scores, because explaining data is its sole objective. 
BP-sLDA is consistently poor, having per-token likelihoods about >1.0 nats higher than others. These results show that the solely discriminative approach of BP-sLDA cannot explain the data well.
In contrast, our PC-sLDA can capture essential data properties while still predicting labels accurately.

\textbf{PC-sLDA's learned topic-word probabilities $\phi$ are interpretable for the prediction task.}
We emphasize that our PC training estimates
topic-word parameters $\phi$ that are distinct from unsupervised training and more appropriate for the label prediction task. Fig.~\ref{fig:results_psych} shows that PC-sLDA initialized from Gibbs indeed causes an original Gibbs topic to significantly evolve its regression weight $\eta_k$ and associated top words.
The original Gibbs topic covers routine outpatient preventative care and vaccination. The evolved PC-sLDA topic prefers long-duration primary care encounters focused on behavior change (``counseling'').
With clinical collaborators, we hypothesize
that this more focused topic
leads to a positive $\eta_k$ value because the drug in question (citalopram/Celexa) is often a treatment of choice for patients with uncomplicated MDD diagnosed and treated in primary care.

\section{Conclusion}

We have presented a new training objective for topic models that can effectively incorporate supervised labels to improve parameter training, even when the model is misspecified. Future work can explore this same PC objective with improved models for observational health records that account for important factors such as 
patient demographics, temporal evolution, or causality.


\newpage
{\small
\linespread{1}
\bibliography{macros_for_journal_names,references}

\begin{thebibliography}{24}
\providecommand{\natexlab}[1]{#1}
\providecommand{\url}[1]{\texttt{#1}}
\expandafter\ifx\csname urlstyle\endcsname\relax
  \providecommand{\doi}[1]{doi: #1}\else
  \providecommand{\doi}{doi: \begingroup \urlstyle{rm}\Url}\fi

\bibitem[Abadi et~al.(2015)Abadi, Agarwal, Barham,
  et~al.]{tensorflow2015whitepaper}
M.~Abadi, A.~Agarwal, P.~Barham, et~al.
\newblock {TensorFlow}: Large-scale machine learning on heterogeneous systems,
  2015.
\newblock Software available from tensorflow.org.

\bibitem[Blei et~al.(2003)Blei, Ng, and Jordan]{blei2003lda}
D.~M. Blei, A.~Y. Ng, and M.~I. Jordan.
\newblock Latent {D}irichlet allocation.
\newblock \emph{Journal of Machine Learning Research}, 3:\penalty0 993--1022,
  2003.

\bibitem[Chen et~al.(2015)Chen, He, Shen, Xiao, He, Gao, Song, and
  Deng]{chen2015bplda}
J.~Chen, J.~He, Y.~Shen, L.~Xiao, X.~He, J.~Gao, X.~Song, and L.~Deng.
\newblock End-to-end learning of {LDA} by mirror-descent back propagation over
  a deep architecture.
\newblock In \emph{Neural Information Processing Systems}, 2015.

\bibitem[Ganchev et~al.(2010)Ganchev, Gra\c{c}a, Gillenwater, and
  Taskar]{ganchev2010posteriorconstraints}
K.~Ganchev, J.~Gra\c{c}a, J.~Gillenwater, and B.~Taskar.
\newblock Posterior regularization for structured latent variable models.
\newblock \emph{Journal of Machine Learning Research}, 11:\penalty0 2001--2049,
  Aug. 2010.

\bibitem[Ghassemi et~al.(2014)Ghassemi, Naumann, Doshi-Velez, Brimmer, Joshi,
  Rumshisky, and Szolovits]{ghassemi2014unfolding}
M.~Ghassemi, T.~Naumann, F.~Doshi-Velez, N.~Brimmer, R.~Joshi, A.~Rumshisky,
  and P.~Szolovits.
\newblock Unfolding physiological state: mortality modelling in intensive care
  units.
\newblock In \emph{ACM SIGKDD International Conference on Knowledge Discovery
  and Data Mining}. ACM, 2014.

\bibitem[Gra\c{c}a et~al.(2008)Gra\c{c}a, Ganchev, and
  Taskar]{gracca2008posteriorconstraints}
J.~Gra\c{c}a, K.~Ganchev, and B.~Taskar.
\newblock Expectation maximization and posterior constraints.
\newblock In \emph{Neural Information Processing Systems}, 2008.

\bibitem[Griffiths and Steyvers(2004)]{griffiths:2004:fst}
T.~L. Griffiths and M.~Steyvers.
\newblock Finding scientific topics.
\newblock \emph{Proceedings of the National Academy of Sciences}, 2004.

\bibitem[Halpern et~al.(2012)Halpern, Horng, Nathanson, Shapiro, and
  Sontag]{halpern2012comparison}
Y.~Halpern, S.~Horng, L.~A. Nathanson, N.~I. Shapiro, and D.~Sontag.
\newblock A comparison of dimensionality reduction techniques for unstructured
  clinical text.
\newblock In \emph{{ICML} workshop on clinical data analysis}, 2012.

\bibitem[Hughes et~al.(2017)Hughes, Weiner, Hope, McCoy, Perlis, Sudderth, and
  Doshi-Velez]{hughes2017pc}
M.~C. Hughes, L.~Weiner, G.~Hope, T.~H. McCoy, R.~H. Perlis, E.~B. Sudderth,
  and F.~Doshi-Velez.
\newblock Prediction-constrained training for semi-supervised mixture and topic
  models.
\newblock \emph{arXiv preprint 1707.07341}, 2017.
\newblock URL \url{https://arxiv.org/pdf/1707.07341.pdf}.

\bibitem[Kivinen and Warmuth(1997)]{kivinen1997exponentiated_gradient}
J.~Kivinen and M.~K. Warmuth.
\newblock Exponentiated gradient versus gradient descent for linear predictors.
\newblock \emph{Information and Computation}, 132\penalty0 (1):\penalty0 1--63,
  1997.

\bibitem[Lacoste-Julien et~al.(2009)Lacoste-Julien, Sha, and
  Jordan]{lacoste2009disclda}
S.~Lacoste-Julien, F.~Sha, and M.~I. Jordan.
\newblock Disc{L}{D}{A}: Discriminative learning for dimensionality reduction
  and classification.
\newblock In \emph{Neural Information Processing Systems}, 2009.

\bibitem[MacKay(1998)]{mackay1998basis}
D.~J.~C. MacKay.
\newblock Choice of basis for {L}aplace approximation.
\newblock \emph{Machine Learning}, 33\penalty0 (1), 1998.

\bibitem[Maclaurin et~al.(2015)Maclaurin, Duvenaud, Johnson, and
  Adams]{maclaurin2015autograd}
D.~Maclaurin, D.~Duvenaud, M.~Johnson, and R.~Adams.
\newblock Autograd: Reverse-mode differentiation of native python.
\newblock \url{http://github. com/HIPS/autograd}, 2015.

\bibitem[Mc{A}uliffe and Blei(2008)]{blei2007sLDA}
J.~D. Mc{A}uliffe and D.~M. Blei.
\newblock Supervised topic models.
\newblock In \emph{Neural Information Processing Systems}, pages 121--128,
  2008.

\bibitem[Paul and Dredze(2014)]{paul2014discovering}
M.~J. Paul and M.~Dredze.
\newblock Discovering health topics in social media using topic models.
\newblock \emph{PLoS One}, 9\penalty0 (8):\penalty0 e103408, 2014.

\bibitem[Ren et~al.(2017)Ren, Wang, and Zhu]{ren2017spectral}
Y.~Ren, Y.~Wang, and J.~Zhu.
\newblock Spectral learning for supervised topic models.
\newblock \emph{IEEE Transactions on Pattern Analysis and Machine
  Intelligence}, 2017.

\bibitem[Sontag and Roy(2011)]{sontag2011complexityoflda}
D.~Sontag and D.~Roy.
\newblock Complexity of inference in latent dirichlet allocation.
\newblock In \emph{Neural Information Processing Systems}, 2011.

\bibitem[Taddy(2012)]{taddy2012topicmodelmapestimation}
M.~Taddy.
\newblock On estimation and selection for topic models.
\newblock In \emph{Artificial Intelligence and Statistics}, 2012.

\bibitem[Wang et~al.(2009)Wang, Blei, and Li]{wang2009simultaneous}
C.~Wang, D.~Blei, and F.-F. Li.
\newblock Simultaneous image classification and annotation.
\newblock In \emph{IEEE Conf. on Computer Vision and Pattern Recognition},
  2009.

\bibitem[Wang and Zhu(2014)]{wang2014spectral}
Y.~Wang and J.~Zhu.
\newblock Spectral methods for supervised topic models.
\newblock In \emph{Advances in Neural Information Processing Systems}, pages
  1511--1519, 2014.

\bibitem[Zhang and Kjellström(2014)]{zhang2014howToSuperviseTopicModels}
C.~Zhang and H.~Kjellström.
\newblock How to supervise topic models.
\newblock In \emph{ECCV Workshop on Graphical Models in Computer Vision}, 2014.

\bibitem[Zhu et~al.(2012)Zhu, Ahmed, and Xing]{zhu2012medlda}
J.~Zhu, A.~Ahmed, and E.~P. Xing.
\newblock Med{L}{D}{A}: maximum margin supervised topic models.
\newblock \emph{The Journal of Machine Learning Research}, 13\penalty0
  (1):\penalty0 2237--2278, 2012.

\bibitem[Zhu et~al.(2013)Zhu, Chen, Perkins, and Zhang]{zhu2013gibbsmaxmargin}
J.~Zhu, N.~Chen, H.~Perkins, and B.~Zhang.
\newblock Gibbs max-margin topic models with fast sampling algorithms.
\newblock In \emph{International Conference on Machine Learning}, 2013.

\bibitem[Zhu et~al.(2014)Zhu, Chen, and Xing]{zhu2014regbayes}
J.~Zhu, N.~Chen, and E.~P. Xing.
\newblock Bayesian inference with posterior regularization and applications to
  infinite latent svms.
\newblock \emph{Journal of Machine Learning Research}, 15\penalty0
  (1):\penalty0 1799--1847, 2014.

\end{thebibliography}
}

\begin{appendix}
\counterwithin{figure}{section}
\section{EHR dataset description}

We study a cohort of hundreds of thousands patients drawn from two large academic medical centers and their affiliated outpatient networks over a period of several years.
Each patient has at least one ICD9 diagnostic code for major depressive
disorder (ICD9s 296.2x or 3x or 311, or ICD10 equivalent).
Each included patient had an identified successful treatment
which included one of 25 possible common anti-depressants marked as ``primary'' treatments for major depressive disorder by clinical collaborators. 
We labeled an interval of a patient's record ``successful'' if 
all prescription events in the interval used the same subset of primary drugs, the interval lasted at least 90 days, and encounters occurred at least every 13 months. 
Applying this criteria, we identified 64431 patients who met our definition of success. For each patient, we extracted a bag-of-codewords $x_d$ of 5126 possible codewords (representing medical history before any successful treatment) and binary label vector $y_d$, marking which of 11 prevalent anti-depressants (if any) were used in known successful treatment.

\paragraph{Extracting data $x_d$.}
For each patient with known successful treatment, we build a data vector $x_d$ to summarize all facts known about the patient in the EHR before any successful treatment was given.
Thus, we must confine our records to the interval from the patient's first encounter to the last encounter before any of the drugs on his or her successful list were first prescribed.
To summarize this patient's interval of ``pre-successful treatment'', 
we built a sparse count vector of all procedures, diagnoses, labs, and medications from the EHR which fit within the interval (22,000 possible codewords).
By definition, none of the anti-depressant medications on the patient's eventual success list appear in $x_d$.
To simplify, we reduced this to a final vocabulary of 5126 codewords that occurred in at least 1000 distinct patients.
We discard any patients with fewer than 2 tokens in $x_d$ (little to no history).

\paragraph{Extracting labels $y_d$.}
Among the 25 primary drugs, we identified a smaller set of 11 anti-depressants which were used in ``successful treatment'' for at least 1000 patients. The remaining 15 primary drugs did not occur commonly enough that we could accurately access prediction quality (build large enough heldout sets). Our chosen list of drugs to predict are:
nortriptyline, amitriptyline, bupropion, fluoxetine, sertraline,
paroxetine, venlafaxine, mirtazapine, citalopram, escitalopram, and duloxetine. Because these drugs can be given in combination, this is a multiple binary label problem. Future work could look into structured prediction tasks.
\section{Toy Bars Case Study}

To study tradeoffs between models of $p(x)$ and $p(y|x)$, 
we built a toy dataset that is deliberately \emph{misspecified}: 
neither the unsupervised LDA maximum likelihood solution nor the standard sLDA joint likelihood solution
performs much better than chance at label prediction.
We look at 500 training documents, each with $V=9$ possible vocabulary words that can be arranged in a 3-by-3 grid to indicate some bar-like co-occurrence structure.
Each binary label $y_d$ is unrelated to the bar structure, but is unambiguously indicated by the top-left word.
We visualize some documents and their associated labels in the top row of Fig.~\ref{fig:results_toy_haystack}.

We compare our proposed PC sLDA training procedure with sevral competitors (MED sLDA \citep{zhu2012medlda}, Gibbs unsupervised LDA \citep{griffiths:2004:fst}, BP sLDA \citep{chen2015bplda}, etc.). 
We also include a method that maximizes the joint likelihood $\log p(x, y)$ with different amounts of label replication. We call this method ML-sLDA with replication value $\lambda$. The special case of $\lambda = 1$ is standard sLDA, while the case of $\lambda \gg 1$ is known as Power sLDA \citep{zhang2014howToSuperviseTopicModels}.

Each algorithm is run to convergence on training data, and its best parameters -- topic-word probabilities $\phi$ and regression weights $\eta$
-- are chosen to minimize method-specific training loss.
Each method's best solution is then located on a 2-dimensional fitness landscape: the x-axis is negative log likelihood of data $x$ averaged per token (lower is better) and y-axis is the negative log likelihood of labels $y$ averaged per document (lower is better). These averages are computed on the \emph{training} set.
We show these fitness scores under two possible modes for estimating each document-topic vector $\pi_d$.
\emph{Train mode} computes the joint likelihood MAP estimate $\max_{\pi_d} \log p(\pi_d | x_d, y_d, \phi, \eta, \alpha)$.
\emph{Predict mode} computes the data-only MAP estimate $\max_{\pi_d} \log p(\pi_d | x_d, \phi, \alpha)$.
This distinction highlights the key difference between PC-sLDA with high $\lambda$, which deliberately trains parameters $\phi, \eta$ to be good at prediction, and alternatives like maximum likelihood with label replication (ML with $\lambda > 1$), which trains models that do well in training mode but fail miserably in a predictive setting (even on the training set). We further see that methods that purely optimize label prediction such as BP-sLDA achieve reasonable prediction scores but \emph{terrible} data likelihood scores.

The visualized parameters show an important trend: 
Our PC-sLDA with $\lambda \geq 10$ is the only method to use just one topic to explain the signal word. 
Thus, it is the only method to reach the sweet spot of good $y|x$ predictions and good $x$ explanations.
Gibbs sensibly finds 4 bars and places the signal word slightly in each one, as does MED-sLDA and Power sLDA. 

\definecolor{gibbs_color}{HTML}{9b59b6}
\definecolor{bpslda_color}{HTML}{1b7837}
\definecolor{pcslda100_color}{HTML}{bc141a}
\definecolor{mlslda100_color}{HTML}{2070b4}
\definecolor{medslda_color}{HTML}{bced91}
\definecolor{pcslda1_color}{HTML}{Fc8a6a}

\begin{figure*}[!h]
\begin{tabular}{c}
\begin{minipage}{3cm}
positive docs:
\end{minipage}%
\begin{minipage}{0.75\textwidth}
\includegraphics[width=\textwidth]{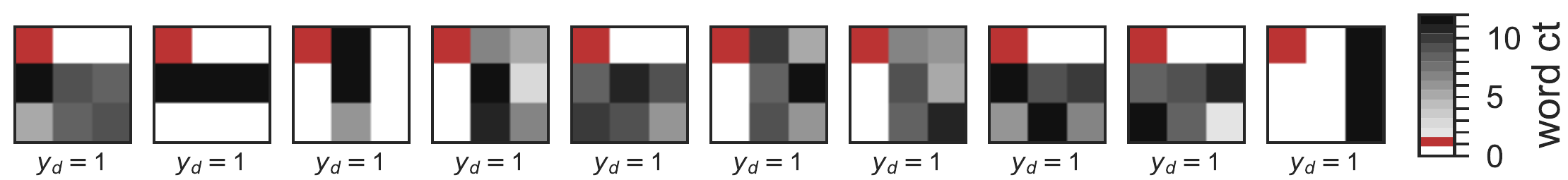} 
\end{minipage}
\\
\begin{minipage}{3cm}
negative docs:
\end{minipage}%
\begin{minipage}{0.75\textwidth}
\includegraphics[width=\textwidth]{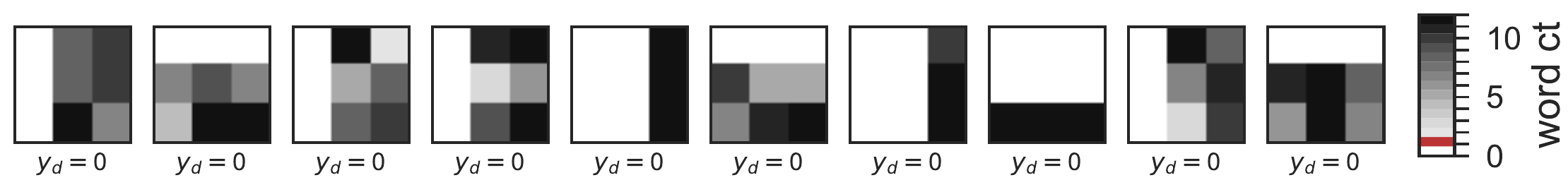} 
\end{minipage}
\\~\\~
\begin{tabular}{c|c}
\textbf{\textcolor{mlslda100_color}{ML sLDA $\lambda=100$ (aka Power sLDA)}}
&
\textbf{\textcolor{pcslda100_color}{PC sLDA $\lambda=100$}}
\\
\includegraphics[width=.42\textwidth]{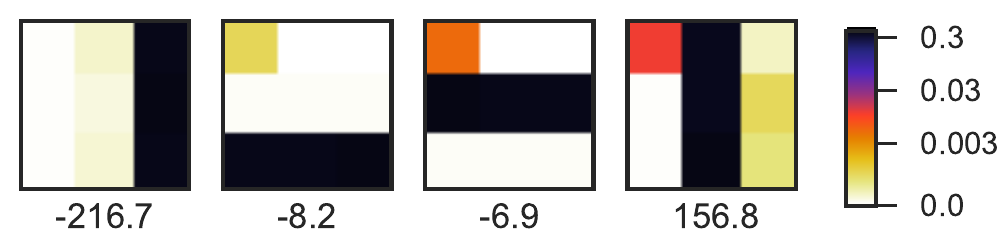} 
&
\includegraphics[width=.42\textwidth]{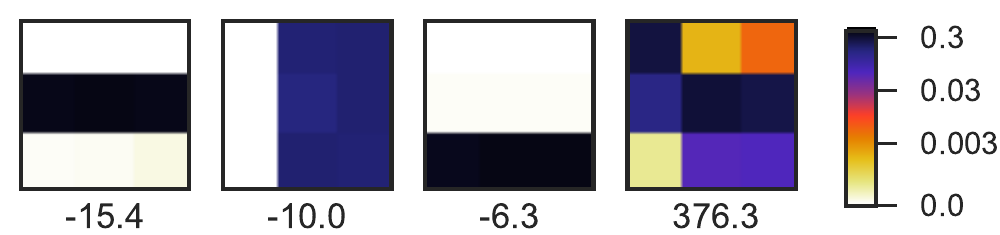} 
\\
\textbf{\textcolor{gibbs_color}{Gibbs LDA}}
&
\textbf{\textcolor{pcslda1_color}{PC sLDA $\lambda=1$}}
\\
\includegraphics[width=0.42\textwidth]{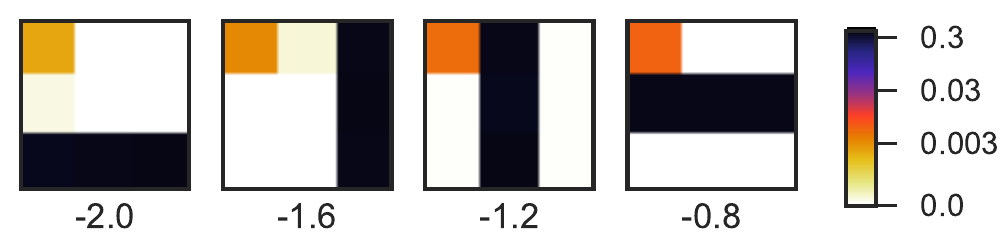} 
&
\includegraphics[width=0.42\textwidth]{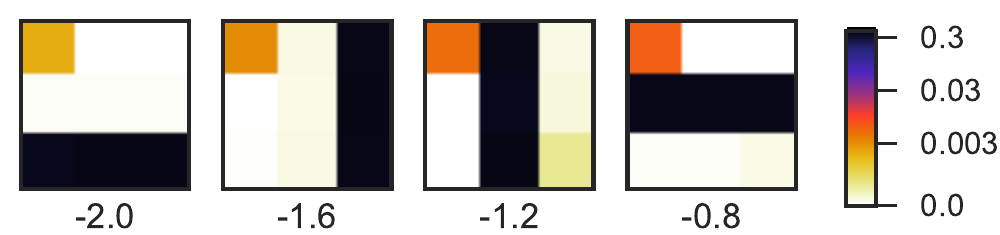} 
\\
\textbf{\textcolor{medslda_color}{MED sLDA}}
&
\textbf{\textcolor{bpslda_color}{BP sLDA}}
\\
\includegraphics[width=0.42\textwidth]{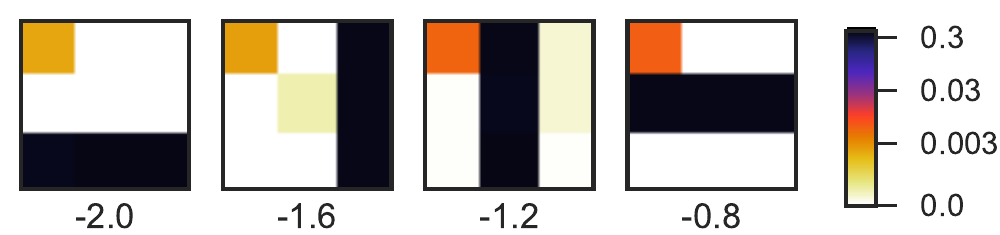} 
&
\includegraphics[width=0.42\textwidth]{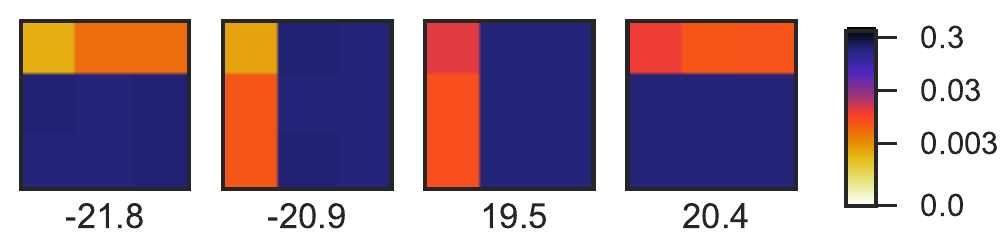} 
\end{tabular}
\\~\\~
\begin{minipage}{0.7\textwidth}
\includegraphics[width=\textwidth]{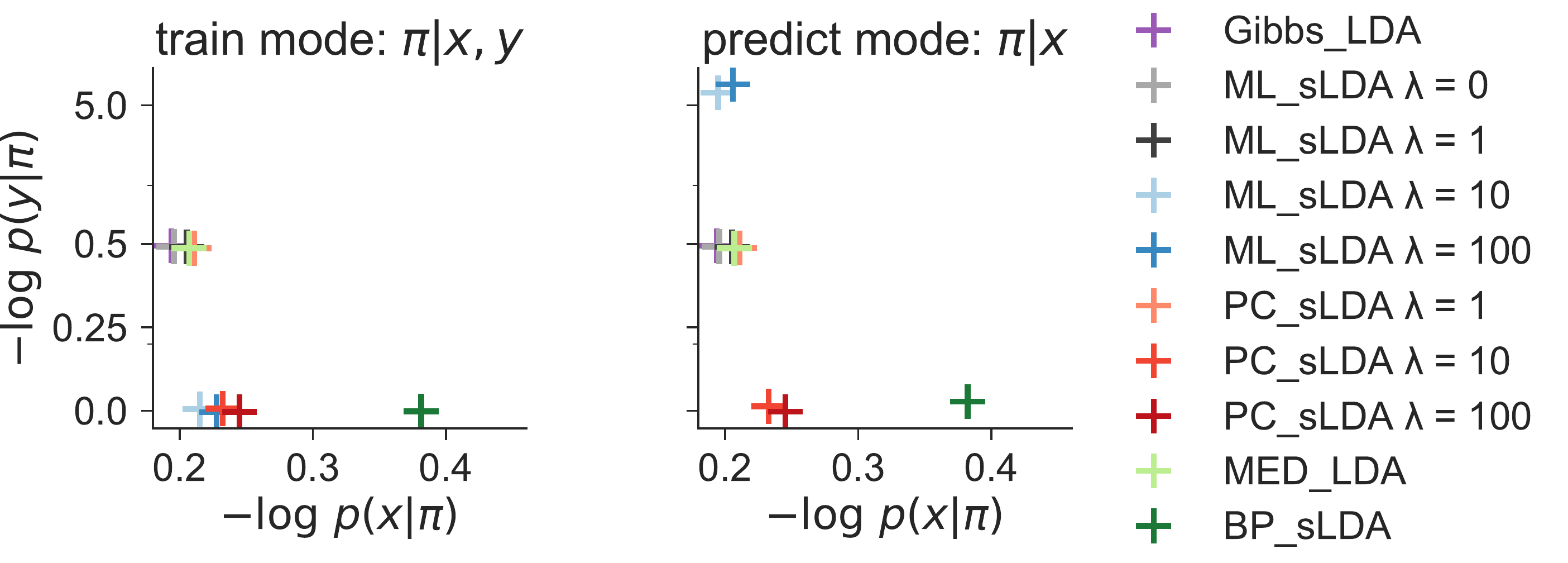}
\end{minipage}%
\end{tabular}
\vspace*{-5pt}
\caption{
$3\times3$ bars task: advantages of PC training under misspecification.
We compare several training procedures
to see which can simultaneously model the bar-like co-occurrence structure while making accurate binary label predictions.
\emph{Top rows:}
Example labeled training documents: the 3$\times$3 heatmap shows the word count vector $x_d$, and the caption indicates the label $y_d$.
Colormap chosen to highlight the top-left-corner symbol that, when it appears just once, perfectly signals the document belongs to the positive class ($y_d=1$). Remaining vocabulary symbols in each document are drawn from one or two of $4$ possible horizontal and vertical ``bar'' topics. These symbols, when non-zero, have much higher counts than the top-left signal word.
\emph{Middle rows:} Visualization of the topic-word probabilities for the best $K=4$ topic model trained by each method. Colormap has a \emph{logarithmic scale} to show how the rare signal word is explained.
\emph{Bottom row:} Location of each method's estimated parameters on the fitness landscape where x-axis is generative model training loss, and y-axis is prediction task loss. The lower left corner is the ideal position.
}
\label{fig:results_toy_haystack}
\end{figure*}

\end{appendix}

\end{document}